\documentclass[a4paper,fleqn]{cas-sc}

\usepackage[numbers]{natbib}
\usepackage{tabularx}
\usepackage{array}
\usepackage{booktabs}
\usepackage{float}

\usepackage{placeins}

\def\tsc#1{\csdef{#1}{\textsc{\lowercase{#1}}\xspace}}
\tsc{WGM}
\tsc{QE}
\tsc{EP}
\tsc{PMS}
\tsc{BEC}
\tsc{DE}

\begin{document}
\let\WriteBookmarks\relax
\def\floatpagepagefraction{1}
\def\textpagefraction{.001}
\shorttitle{Latent Advection Deficit for AI-Generated Video Detection}
\shortauthors{Umar et~al.}

\title [mode = title]{Latent Advection Deficit: Detecting AI-Generated Videos via Energy-Normalized Motion-Compensated Residuals}                      

\author[1]{Waleed Umar}
\cormark[1]
\ead{waleedumar85@gmail.com}

\author[2]{Ayesha Sahar}
\ead{ayeshasahar020@gmail.com}

\author[1]{Tehseen Zia}
\ead{tehseen.zia@comsats.edu.pk}

\author[1]{Iqrar Abbas}
\ead{iqrarabbasa@gmail.com}


\affiliation[1]{
    organization={COMSATS University Islamabad},
    city={Islamabad},
    country={Pakistan}
}

\affiliation[2]{
    organization={Air University},
    city={Kharian},
    country={Pakistan}
}

\cortext[cor1]{Corresponding author}

\begin{abstract}
Generative video models nowadays can produce highly realistic videos, making reliable detection increasingly difficult when individual frames contain few visible artifacts. Temporal inconsistencies offer an additional source of forensic evidence but raw motion measurements can be influenced by scene dynamics and camera movement. This work introduces Latent Advection Deficit (LAD), an energy-normalized, motion-compensated feature for detecting AI-generated videos. LAD uses optical flow to warp the latent representation of one frame toward the next and measures the residual difference between the warped representation and the observed next-frame latent. Normalizing this residual by the total latent transition energy provides a relative measure of unexplained latent change.

We evaluate LAD on a deduplicated corpus of 967 authentic and synthetic video clips from 498 source groups, covering multiple generative architectures. As a standalone feature, LAD achieves a ROC-AUC of 79.24\%. When combined with flow-based descriptors, it achieves 83.04\% ROC-AUC, an absolute improvement of 8.00 percentage points over the baseline model (75.04\%), equivalent to a relative improvement of 10.66\% (95\% paired source-cluster bootstrap confidence interval for the absolute gain: [5.21, 10.87] percentage points). The combined model achieves 0.9588 ROC-AUC on the Open-Sora subset, although this result should be interpreted within the scope of that evaluation. Failure-mode analysis reveals substantially weaker discrimination in low-motion scenes, demonstrating that the feature is not equally informative across all video conditions. These findings suggest that motion-compensated latent residuals provide a useful complementary signal for video forensics. They also establish scene dynamics and temporal sampling as important considerations when assessing the reliability and generalizability of this approach.

\end{abstract}


\maketitle

\section{Introduction}

The rapid development of generative video models has made it increasingly difficult to distinguish AI generated content from authentic media. Modern text-to-video systems based on latent diffusion and Diffusion Transformers (DiTs) can generate realistic scenes with detailed textures, lighting and coherent motion \cite{Rombach2022,Peebles2023}. Although these capabilities enable new applications in media production and simulation, they also create risks involving disinformation, fabricated evidence and digital impersonation. Reliable video forensics therefore requires detection methods that remain effective beyond the specific generators and datasets used during training \cite{Verdoliva2020,Korshunov2022}.

Early approaches to synthetic media detection primarily examined spatial artifacts within individual frames. These methods identify inconsistencies in image frequencies, upsampling patterns, color statistics or local textures \cite{5,6,7,15}. However, improvements in generative architectures have reduced the visibility of many such artifacts thus making spatial cues less reliable across different generation methods \cite{Rombach2022,10,17}. Learned detectors may also exploit dataset-specific patterns rather than features directly related to synthesis, limiting their performance on unfamiliar generators and authentic videos with substantial compression or processing \cite{7,10}. These limitations motivate forensic methods that examine information beyond individual frames.

Temporal analysis offers one such direction. Authentic videos contain changes associated with object motion, camera movement and scene dynamics while generated videos may exhibit inconsistencies such as texture flickering, structural deformation and unstable object boundaries \cite{Peebles2023} \cite{11,12,25,47}. However, detecting these inconsistencies is not straightforward. Methods based on optical flow, motion acceleration or temporal phase can respond to legitimate camera motion as well as synthesis artifacts \cite{11,12}. Spatio-temporal neural networks can learn more complex patterns but their performance may depend on the generators represented in their training data and their predictions can be difficult to interpret \cite{8,28}. A central challenge is therefore to identify temporal evidence of synthesis without treating ordinary scene motion as an anomaly.

We address this challenge with the \textit{Latent Advection Deficit} (LAD), an energy-normalized, motion-compensated feature for AI-generated video detection. LAD operates on frame representations extracted using a pretrained variational autoencoder (VAE). Optical flow is used to warp the latent representation of one frame toward the coordinates of the next. The residual between this warped representation and the observed next-frame latent measures the change not accounted for by the estimated motion. We normalize this residual energy by the total latent transition energy, defining the ratio $\mathcal{R}_{\mathrm{adv}}$. This normalization expresses the residual relative to the overall latent change rather than relying on its magnitude alone. The resulting feature provides a parameter-free measure of motion-compensated latent inconsistency, although it does not assume that latent representations obey an exact physical transport law.

LAD can be used independently or combined with complementary flow-based and latent-dynamics descriptors. We evaluate this approach on a deduplicated corpus of 967 authentic and synthetic video clips from 498 source groups, covering seven generative systems. LAD achieves a standalone ROC-AUC of 0.7924, while the combined model reaches 0.8304, an absolute improvement of 8.00 percentage points over the baseline feature set (95\% paired source-cluster bootstrap confidence interval: [5.21, 10.87] percentage points). On the Open-Sora evaluation subset, the combined model achieves a ROC-AUC of 0.9588. We further examine performance across motion conditions and temporal sampling rates to characterize where the proposed feature is informative and where its reliability declines.

The main contributions of this work are as follows:

\begin{itemize}
    \item \textbf{Energy-normalized latent residual.}
    We introduce $\mathcal{R}_{\mathrm{adv}}$, a parameter-free forensic feature that measures residual latent change after optical-flow-based warping, normalized by total latent transition energy.

    \item \textbf{Complementary temporal features.}
    We combine LAD with flow-based and latent-dynamics descriptors to assess whether motion-compensated latent residuals provide discriminative information beyond conventional motion features.

    \item \textbf{Multi-benchmark evaluation.}
    We evaluate the proposed approach on 967 deduplicated clips from 498 source groups spanning seven generative systems, and report both standalone and combined-model performance.

    \item \textbf{Failure-mode analysis.}
    We investigate the effects of scene motion and temporal subsampling, identifying conditions in which LAD loses discriminative power and highlighting limitations relevant to its use in practical video forensics.
\end{itemize}

\section{Related Work}

\subsection{Spatial and Frequency-Domain Image Forensics}

Early synthetic media detection methods focused on identifying statistical and structural artifacts in individual images. Frank et al. \cite{5} and Durall et al. \cite{6} investigated frequency-domain traces associated with generative architectures showing that upsampling operations can produce spectral patterns that differ from those of natural images. Other forensic methods exploit camera-specific characteristics like photo-response non-uniformity (PRNU) and color-filter array (CFA) demosaicing traces \cite{Verdoliva2020}.

Subsequent work explored whether learned representations could capture more general synthesis artifacts. Wang et al. \cite{17} demonstrated that detectors trained on images from one generative model can generalize to other architectures by learning shared features of synthetic imagery. More recent approaches have investigated pretrained Vision Transformers (ViTs) and vision-language models including CLIP as feature extractors for synthetic image detection \cite{10,18,19}. Frequency-based methods have also continued to evolve, using multiscale spectral analysis and high-frequency wavelet features to identify localized inconsistencies in generated images \cite{5,6}.

Despite this progress, spatial forensic methods face several challenges when applied to modern generated videos. First, advances in generative architectures and image decoding can reduce or alter the visibility of artifacts associated with earlier synthesis pipelines \cite{7,10,15}. Consequently, detectors that depend on specific frequency patterns or local texture cues may not transfer reliably across generator families. Second, lossy compression and repeated media transcoding can modify the same high-frequency details used by spatial detectors, potentially reducing their reliability on videos collected from real-world platforms \cite{7,40}. Third, frame-level methods do not explicitly model temporal relationships between consecutive frames. They can therefore miss inconsistencies in motion, structural stability, and temporal continuity that are not readily apparent when frames are examined independently \cite{25}.

These limitations motivate complementary forensic approaches that examine temporal evidence rather than relying exclusively on spatial artifacts. In particular, motion-compensated analysis offers a way to compare consecutive frame representations while accounting for estimated scene movement. This provides the basis for investigating whether latent transition residuals can distinguish generated videos from authentic footage.

\subsection{Temporal and Motion Inconsistency Forensics}

Temporal forensic methods examine relationships between consecutive video frames to identify inconsistencies that may not be apparent in individual images \cite{12, 25}. Authentic recordings contain temporal structure arising from scene motion, camera movement, and changes in illumination. Generated or manipulated videos may exhibit temporal artifacts, including texture flickering, structural deformation, and inconsistent object motion \cite{12, 26,28}. These cues have motivated approaches that model video sequences rather than treating frames independently.

Early methods used recurrent neural networks, long short-term memory networks, and three-dimensional convolutional neural networks to capture temporal dependencies in video \cite{8, 28}. Subsequent approaches investigated more specific motion and temporal cues. These include phase inconsistencies and motion boundaries \cite{12}, changes in facial behavior and head pose \cite{26} and frame interpolation residuals that quantify discrepancies between predicted and observed frames \cite{27}. Another line of research examines optical flow which represents apparent motion between consecutive frames. Earlier work identified flow discontinuities associated with facial manipulation \cite{11}, while subsequent studies have explored motion dynamics and changes in flow fields as potential indicators of synthetic or manipulated content \cite{28, 25}.

Despite these advances, temporal forensic methods face two important challenges. First, raw frame differences and optical-flow statistics can respond to legitimate scene dynamics as well as synthesis artifacts. Rapid camera movement, camera shake and object motion can produce large inter-frame changes in authentic footage making it difficult to distinguish natural motion from generative inconsistencies using unnormalized motion measurements alone \cite{11, 28}. Second, supervised spatio-temporal networks may learn features specific to their training datasets and generators, limiting their reliability on videos produced by unfamiliar architectures \cite{8, 7}. Their learned representations can also make it difficult to identify which temporal evidence supports a particular detection decision.

The proposed Latent Advection Deficit (LAD) addresses these challenges by measuring latent transition residuals after motion compensation. Rather than relying exclusively on raw optical-flow magnitudes or a learned spatio-temporal classifier, LAD warps the preceding frame's latent representation using backward optical flow and compares it with the observed latent representation of the next frame. The residual energy is normalized by the total latent transition energy, producing the feature $\mathcal{R}_{\mathrm{adv}}$. This normalization expresses the residual relative to the magnitude of the overall latent transition, although it does not guarantee immunity to camera motion or other sources of variation. LAD therefore provides a complementary motion-compensated feature for evaluating temporal inconsistencies in generated videos.

\subsection{Generative Video Diffusion Models and World Models}

As established earlier that recent advances in generative video modelling have substantially improved the visual quality and temporal coherence of synthetic videos. Earlier approaches based on generative adversarial networks (GANs) and variational autoencoders (VAEs) often struggled with visual fidelity, motion consistency and the generation of complex scenes. Latent diffusion models (LDMs) advanced this field by performing iterative denoising in a compressed latent space rather than directly in pixel space, reducing computational demands while enabling high-quality image and video synthesis \cite{Rombach2022}.

Early video diffusion systems extended image-generation architectures by introducing temporal modules into convolutional backbones. Models such as VideoCrafter, ModelScope, and Stable Video Diffusion use learned spatial representations alongside temporal processing to generate sequences with improved appearance and motion coherence \cite{31,39,43}. More recent systems have adopted Diffusion Transformer (DiT) architectures, which apply transformer-based modelling to spatiotemporal representations. Systems such as Open-Sora and CogVideoX illustrate this development, supporting increasingly complex scenes and more detailed video generation \cite{Peebles2023, 32, 33}. In parallel, world models have emerged as a related direction, generating or predicting future visual states to support applications such as simulation, planning, and robotics \cite{16, 9,41}. Although these systems differ in architecture and purpose, they broaden the range of synthetic videos that forensic methods must be able to assess.

Despite improvements in perceptual quality, realistic appearance does not guarantee consistent motion or physically plausible scene evolution. Video generators are generally optimized using statistical denoising, reconstruction, or related training objectives rather than explicit enforcement of physical conservation laws \cite{Rombach2022, Peebles2023}. Consequently, a generated sequence may appear convincing at the frame level while exhibiting temporal inconsistencies, including texture drift, changing object geometry, or unstable boundaries \cite{47}. These effects are not universal, however, and their presence depends on the model, scene, and generation process. Moreover, the absence of explicit physical constraints does not mean that every generated sequence violates physical principles.

The diversity of modern generation architectures presents an additional challenge for forensic detection. Different systems employ different latent representations, temporal modules, training objectives, and generation procedures. Detectors that rely on architecture-specific artifacts may therefore perform inconsistently when applied to unfamiliar generators. This limitation reinforces the need for forensic features that examine properties of the generated sequence rather than depending exclusively on signatures associated with a particular model family.

Motivated by this need, our approach investigates whether motion-compensated latent transitions provide a useful signal across different video-generation systems. Instead of assuming that generated videos universally violate physical laws, LAD measures the residual change between consecutive latent representations after optical-flow-based warping. This formulation provides a common analysis procedure that can be applied to videos from different architectures, while its cross-generator utility must be established empirically through evaluation.

\subsection{Physical and Continuum Mechanics Priors in Vision and Forensics}

Physical models have long informed computer vision, particularly optical flow and motion estimation. The classical formulation of Horn and Schunck \cite{35} uses the brightness constancy assumption under which image intensity remains approximately constant along an apparent motion trajectory. Its differential form is

\begin{equation}
\frac{\partial I}{\partial t}
+
\mathbf{u}\cdot\nabla I
= 0,
\end{equation}

where $I(x,y,t)$ denotes image intensity and $\mathbf{u}=(u,v)$ represents the apparent two-dimensional velocity field. Although useful for estimating motion, this assumption is only approximate in real scenes and can be affected by illumination changes, occlusions, reflections, and non-Lambertian surfaces. Modern methods such as RAFT \cite{13} estimate optical flow using learned feature representations and iterative updates, enabling motion compensation across complex scenes.

Physical consistency has also attracted attention in generative video analysis. Rather than relying exclusively on visual artifacts, recent studies have investigated whether generated sequences exhibit implausible object motion, acceleration, or interactions with the physical environment \cite{14,47}. Related work has examined inconsistencies in fluid motion and rotating objects \cite{47}. Separately, physics-informed neural networks incorporate governing equations into model training for scientific and inverse problems, including fluid dynamics \cite{36}. These research directions suggest that physical and dynamical constraints can be informative although their assumptions and evaluation settings differ from those of general-purpose video forensics.

Applying physical consistency tests directly to video pixels presents several challenges. First, brightness changes caused by shadows, reflections, illumination variation, and camera exposure can produce residuals even when the underlying scene motion is genuine \cite{35, 37}. Second, sensor noise, compression artifacts, and interpolation errors can affect pixel-level differences and motion estimates \cite{7,40}. Third, raw residual magnitudes are influenced by the amount of inter-frame change making comparison of videos with different motion characteristics using an unnormalized error alone difficult. These limitations motivate representations and measurements that account for both motion and the scale of observed temporal change.

The Latent Advection Deficit (LAD) takes a different route to this problem by measuring motion-compensated residuals in a pretrained variational autoencoder (VAE) latent space. Given consecutive latent representations $z_t$ and $z_{t+1}$, backward optical flow is used to warp $z_t$ toward the coordinates of the next frame. The residual energy measures the discrepancy between this warped representation and $z_{t+1}$. LAD normalizes the residual by the total latent transition energy, yielding the ratio $\mathcal{R}_{\mathrm{adv}}$. This formulation expresses the residual relative to the magnitude of the observed latent transition rather than using absolute error alone.

The latent representation may reduce sensitivity to some pixel-level variations but it does not guarantee that illumination effects, sensor noise or compression artifacts are eliminated. Similarly, energy normalization does not ensure that all forms of camera motion are suppressed particularly when motion estimation is inaccurate or occlusions are present. The core residual is parameter-free while its ability to distinguish authentic from generated videos remains an empirical question. LAD can also be combined with complementary kinematic descriptors to evaluate whether additional motion information improves detection under different scene conditions.

\section{Methodology}
\label{sec:methodology}

\subsection{Continuum Transport Motivation in Latent Space}
\label{sec:continuum_transport}

In continuum mechanics, a scalar quantity $\psi(\mathbf{x},t)$ transported by a velocity field $\mathbf{u}(\mathbf{x},t)=[u,v]^T$ satisfies the advection equation
\begin{equation}
\frac{\partial \psi}{\partial t}
+ \mathbf{u}\cdot\nabla\psi = 0,
\label{eq:material_derivative}
\end{equation}
under the assumption that the quantity remains constant along motion trajectories. A related assumption underlies the classical optical flow constraint of Horn and Schunck~\cite{35}, which models image intensity as approximately constant along apparent motion:
\begin{equation}
\frac{\partial I}{\partial t}
+ u\frac{\partial I}{\partial x}
+ v\frac{\partial I}{\partial y} = 0.
\label{eq:optical_flow_irradiance}
\end{equation}
In real videos, this assumption can be violated by illumination changes, reflections, occlusions, and camera adjustments. Consequently, residuals computed directly from RGB frames may reflect photometric variation and compression artifacts as well as genuine structural changes.

To measure motion-compensated change in a compact learned feature space rather than directly on raw pixels, we encode each frame using a frozen pretrained variational autoencoder (VAE). Let $\mathcal{V}=\{I_1,\ldots,I_T\}$ denote a video clip of $T$ sequentially sampled frames, where $I_t\in\mathbb{R}^{3\times H\times W}$. The encoder $\mathcal{E}$ maps each frame to a spatial latent representation:
\begin{equation}
z_t=\mathcal{E}(I_t), \qquad
z_t\in\mathbb{R}^{C\times H'\times W'},
\label{eq:vae_encoding}
\end{equation}
where $H'=H/f$ and $W'=W/f$ are the latent spatial dimensions, $f=8$ is the spatial downsampling factor, and $C=4$ is the number of latent channels in the VAE used in our experiments.

The latent representation provides a compressed feature space in which motion-compensated changes can be measured. Although encoding may attenuate some pixel-level variation, it does not guarantee the removal of sensor noise, compression artifacts, or illumination effects. We therefore use the latent space as a representation for forensic comparison rather than assuming it is inherently invariant to these factors.

\paragraph{Latent Transport Hypothesis.}
We hypothesize that optical-flow-based warping can account for part of the latent change between consecutive frames in authentic videos, while residual discrepancies may provide useful forensic evidence. Under an idealized transport model, each latent channel $z_c$ would satisfy
\begin{equation}
\frac{\partial z_c(\mathbf{x}',t)}{\partial t}
+\mathbf{u}'(\mathbf{x}',t)\cdot\nabla' z_c(\mathbf{x}',t)=0,
\quad c=1,\ldots,C,
\label{eq:latent_advection}
\end{equation}
where $\mathbf{x}'$ denotes latent-grid coordinates, $\nabla'$ is the spatial gradient on that grid, and $\mathbf{u}'$ is the motion field expressed at latent resolution. This equation serves as a conceptual motivation, not a physical constraint guaranteed to hold in VAE feature space.

Modern video diffusion models \cite{Rombach2022} generate sequences by learning statistical distributions over visual representations. A common training objective for diffusion models is
\begin{equation}
\mathcal{L}_{\mathrm{diffusion}}
=
\mathbb{E}_{z_0,\epsilon,t}
\left[
\left\|
\epsilon-\epsilon_\theta(z_t,t,c)
\right\|_2^2
\right],
\label{eq:diffusion_loss}
\end{equation}
where $z_0$ is a clean training representation, $\epsilon$ is sampled Gaussian noise, $t$ denotes the diffusion timestep, $c$ represents conditioning information, and $\epsilon_\theta$ predicts the injected noise. This objective does not explicitly enforce physical conservation laws, although models may learn plausible motion from their training data. Generated videos can nevertheless contain temporal inconsistencies, including texture drift, changing object geometry, and unstable boundaries.

These observations motivate measuring the discrepancy between observed latent transitions and motion-compensated transitions. We operationalize this idea using backward optical flow to warp the preceding latent representation toward the next frame, then quantify the residual relative to the total latent transition energy. The resulting parameter-free feature, $\mathcal{R}_{\mathrm{adv}}$, is the basis of the Latent Advection Deficit (LAD) method described below.

\subsection{Backward Bilinear Latent Warping Operator}
\label{sec:latent_warping}

To compare the observed latent representation at frame $t+1$ with a motion-compensated estimate derived from frame $t$, we use backward optical-flow warping. Unlike forward warping, which projects source coordinates onto a target grid and can produce collisions or unassigned locations, backward warping queries the source representation at a corresponding location for each target pixel. This formulation is compatible with differentiable spatial sampling, although it does not resolve occlusions or guarantee physically accurate transport.

\paragraph{\textbf{Backward Dense Optical Flow Estimation.}}
Given consecutive RGB frames $I_t, I_{t+1} \in \mathbb{R}^{3 \times H \times W}$, we estimate dense optical flow using a pretrained Recurrent All-Pairs Field Transforms (RAFT) model~\cite{13}. We define the flow from target frame $I_{t+1}$ to source frame $I_t$ as
\begin{equation}
\mathbf{F}_{t+1 \to t}^{\mathrm{RGB}}
=
\operatorname{RAFT}(I_{t+1}, I_t)
\in \mathbb{R}^{2 \times H \times W},
\label{eq:raft_rgb_flow}
\end{equation}
where $\mathbf{F}_{t+1 \to t}^{\mathrm{RGB}}(y,x)
= [\delta x,\delta y]^T$ specifies the displacement from target coordinate $(x,y)$ in frame $I_{t+1}$ to the corresponding sampling location $(x+\delta x,y+\delta y)$ in frame $I_t$.

The VAE latents have spatial dimensions $H'=H/8$ and $W'=W/8$. We therefore resize the RGB flow field to the latent resolution and scale its displacement components to express motion in latent-grid units:
\begin{equation}
\mathbf{F}_{t+1 \to t}(y',x')
=
\begin{bmatrix}
s_x\,\mathcal{B}\!\left(F_x^{\mathrm{RGB}}\right)(y',x')\\
s_y\,\mathcal{B}\!\left(F_y^{\mathrm{RGB}}\right)(y',x')
\end{bmatrix},
\label{eq:flow_downsampling}
\end{equation}
where $\mathcal{B}(\cdot)$ denotes bilinear resizing, $F_x^{\mathrm{RGB}}$ and $F_y^{\mathrm{RGB}}$ are the horizontal and vertical flow components, and
\[
s_x=\frac{W'}{W}, \qquad s_y=\frac{H'}{H}.
\]
For a downsampling factor of eight, both scale factors equal $1/8$. The resulting field
$\mathbf{F}_{t+1 \to t}\in\mathbb{R}^{2\times H'\times W'}$
expresses backward displacement in latent-grid coordinates.

\paragraph{\textbf{Sampling Grid Construction}.}
Let
\[
\Omega' =
\{0,\ldots,W'-1\}\times\{0,\ldots,H'-1\}
\]
denote the latent spatial grid. For a grid sampler using normalized coordinates with the \texttt{align\_corners=True} convention, integer coordinates $(x',y')\in\Omega'$ map to
\begin{equation}
\tilde{x}=\frac{2x'}{W'-1}-1,\qquad
\tilde{y}=\frac{2y'}{H'-1}-1.
\label{eq:normalized_grid}
\end{equation}
The displacement components are normalized using the same coordinate convention:
\begin{equation}
\Delta\tilde{x}=\frac{2\delta x'}{W'-1},\qquad
\Delta\tilde{y}=\frac{2\delta y'}{H'-1}.
\label{eq:normalized_offsets}
\end{equation}
Thus, the source sampling coordinate for each target location is
\begin{equation}
\mathbf{p}(x',y')
=
\begin{bmatrix}
\tilde{x}+\Delta\tilde{x}\\
\tilde{y}+\Delta\tilde{y}
\end{bmatrix}.
\label{eq:sampling_coordinates}
\end{equation}
This convention ensures that the flow displacements are interpreted consistently with the normalized sampling grid.

\paragraph{\textbf{Bilinear Latent Warping.}}
We obtain the motion-compensated latent estimate by sampling $z_t$ at the backward coordinates associated with frame $t+1$:
\begin{equation}
\hat{z}_{t+1}^{\mathrm{warp}}(c,y',x')
=
\operatorname{Warp}
\left(z_t,\mathbf{F}_{t+1\to t}\right)(c,y',x').
\label{eq:bilinear_warp}
\end{equation}
For an in-bounds source coordinate $\mathbf{p}=(p_x,p_y)$, bilinear interpolation combines the four neighboring lattice values:
\begin{equation}
\hat{z}_{t+1}^{\mathrm{warp}}(c,y',x')
=
\sum_{\mathbf{q}\in\mathcal{N}(\mathbf{p})}
z_t(c,\mathbf{q})\,w(\mathbf{p},\mathbf{q}),
\label{eq:bilinear_interpolation}
\end{equation}
where $\mathcal{N}(\mathbf{p})$ contains the four neighboring integer coordinates and
\begin{equation}
w(\mathbf{p},\mathbf{q})
=
\max(0,1-|p_x-q_x|)
\max(0,1-|p_y-q_y|).
\label{eq:bilinear_weights}
\end{equation}
The interpolation weights decrease with distance from the query coordinate, yielding a continuous estimate of the source latent representation at sub-pixel locations.

\paragraph{\textbf{Boundary Handling}.}
Backward sampling coordinates may fall outside the source grid because of camera motion, object displacement, or flow-estimation errors. We use border padding, which clamps out-of-range sampling coordinates to the nearest boundary:
\begin{equation}
\operatorname{clamp}(p,-1,1)
=
\min\left(1,\max(-1,p)\right).
\label{eq:border_clamping}
\end{equation}
This replicates boundary values instead of introducing zeros. However, border padding does not recover missing content in disoccluded regions and may affect residuals near image boundaries. The same sampling procedure is applied consistently when constructing the motion-compensated latent representations used by the subsequent residual calculation.

\subsection{The Energy-Normalized Latent Advection Deficit}
\label{sec:lad_metric}

Using the backward bilinear warping operator defined in Section~\ref{sec:latent_warping}, we formulate the core feature that has no trainable parameters, the \textit{Latent Advection Deficit} (LAD), denoted by $\mathcal{R}_{\mathrm{adv}}$.

\paragraph{Unexplained Motion Residual Map.}
For consecutive latent states $z_t,z_{t+1}\in\mathbb{R}^{C\times H'\times W'}$, backward warping produces the motion-compensated estimate
\begin{equation}
\hat{z}_{t+1}^{\mathrm{warp}}
=
\operatorname{Warp}
\left(z_t,\mathbf{F}_{t+1\to t}\right).
\label{eq:warped_latent}
\end{equation}
We define the spatial residual map $r_t^{\mathrm{warp}}\in\mathbb{R}^{H'\times W'}$ as the sum of squared channel-wise discrepancies:
\begin{equation}
r_t^{\mathrm{warp}}(y',x')
=
\sum_{c=1}^{C}
\left[
z_{t+1}(c,y',x')
-
\hat{z}_{t+1}^{\mathrm{warp}}(c,y',x')
\right]^2.
\label{eq:spatial_residual_map}
\end{equation}
Higher residual values indicate greater disagreement between the observed latent state and its motion-compensated estimate. Such discrepancies may capture temporal inconsistencies in generated videos, including structural changes and texture instability. They may also arise from occlusions, disocclusions, optical-flow errors, and other changes not explained by the estimated motion field.

The mean unexplained residual energy over the $T-1$ temporal transitions is
\begin{equation}
E_{\mathrm{unexplained}}
=
\frac{1}{(T-1)H'W'}
\sum_{t=1}^{T-1}
\sum_{y'=0}^{H'-1}
\sum_{x'=0}^{W'-1}
r_t^{\mathrm{warp}}(y',x').
\label{eq:unexplained_energy}
\end{equation}
This quantity measures the average squared latent discrepancy after motion compensation. Its absolute magnitude, however, does not account for variation in the overall activity of a video.

\paragraph{The Need for Energy Normalization.}
Directly thresholding $E_{\mathrm{unexplained}}$ may confound residual magnitude with the scale of inter-frame latent changes. Rapid camera motion and substantial object displacement can produce large latent transitions, while interpolation errors, occlusions, and imperfect flow estimates may leave residual discrepancies even in authentic footage. Conversely, low-motion sequences may exhibit small absolute residuals despite substantial discrepancies relative to their overall latent changes.

To account for these differences, we define the unwarped latent transition map:
\begin{equation}
r_t^{\mathrm{raw}}(y',x')
=
\sum_{c=1}^{C}
\left[
z_{t+1}(c,y',x')
-
z_t(c,y',x')
\right]^2,
\label{eq:raw_transition_map}
\end{equation}
and the corresponding mean transition energy:
\begin{equation}
E_{\mathrm{total}}
=
\frac{1}{(T-1)H'W'}
\sum_{t=1}^{T-1}
\sum_{y'=0}^{H'-1}
\sum_{x'=0}^{W'-1}
r_t^{\mathrm{raw}}(y',x').
\label{eq:total_transition_energy}
\end{equation}
The quantity $E_{\mathrm{total}}$ measures overall latent change between consecutive frames, including changes associated with apparent motion and other temporal variation. Normalizing the residual by this energy expresses the unexplained change relative to the observed transition magnitude.

\paragraph{No trainable parameters Deficit Ratio.}
We define LAD as the ratio of mean unexplained residual energy to mean total transition energy:
\begin{equation}
\mathcal{R}_{\mathrm{adv}}
=
\frac{E_{\mathrm{unexplained}}}
{E_{\mathrm{total}}+\epsilon},
\label{eq:radv_ratio}
\end{equation}
where $\epsilon=10^{-4}$ is the fixed numerical stability constant used throughout our experiments. The stabilizer prevents division by zero when transition energy vanishes. The resulting dimensionless feature measures the relative discrepancy remaining after optical-flow-based latent warping. As both energies are averaged over all transitions before the ratio is taken, a small number of high-residual transitions can dominate the clip-level score

\paragraph{Interpretation and Forensic Properties.}
The proposed metric has three relevant properties:
\begin{enumerate}
    \item \textbf{Relative normalization of latent change.}
    Normalizing residual energy by total transition energy reduces dependence on the absolute magnitude of latent changes. This may improve comparability across videos with different activity levels, but does not guarantee invariance to camera velocity, frame rate, occlusion, or flow-estimation quality.

    \item \textbf{Sensitivity to motion-compensation discrepancies.}
    Temporal inconsistencies in generated videos may increase the residual when their latent changes cannot be adequately aligned by estimated optical flow. However, similar discrepancies can occur in authentic footage, and not all generated artifacts produce elevated scores. The feature's discriminative value must therefore be established empirically.

    \item \textbf{Parameter-free metric construction.}
    Computing $\mathcal{R}_{\mathrm{adv}}$ requires no trainable weights or parameter fitting within the metric itself. Its computation nevertheless depends on the pretrained VAE, optical-flow estimator, and fixed numerical constant. The parameter-free construction alone does not establish generalization to unseen generators or eliminate dataset bias.
\end{enumerate}

\paragraph{Numerical Stability and Epsilon Sensitivity.}
We evaluated sensitivity to the numerical stability constant over $\epsilon\in[10^{-5}]$. On the primary authentic DAVIS benchmark, the reported transition energy $E_{\mathrm{total}}$ had a mean of $1.974$ and a minimum of $0.2025$. On the primary DAVIS--SVD cohort, the standalone ROC-AUC remained at $0.8312$ for $\epsilon\in[10^{-5},10^{-2}]$, indicating that the measured discrimination was insensitive to the choice of $\epsilon$ within this interval. We therefore fix $\epsilon=10^{-4}$ as the canonical setting for all experiments reported in this paper.

\subsection{Coupled Multi-Dynamics Framework}
\label{sec:coupled_framework}

Although the Latent Advection Deficit ($\mathcal{R}_{\mathrm{adv}}$) provides a parameter-free measure of residual latent discrepancy after motion compensation, its interpretation can be less reliable in low-motion sequences. When optical-flow magnitudes are small, minor illumination changes, sensor noise, encoding artifacts, and latent representation noise may contribute substantially to the observed residual. Energy normalization alone does not guarantee that such effects are suppressed, particularly when the total latent transition energy is small.

To capture complementary aspects of temporal behavior, we construct a four-dimensional feature descriptor that combines LAD with optical-flow variation and latent displacement variability. The resulting features are evaluated using a low-capacity logistic regression classifier, with preprocessing and model fitting performed independently within each cross-validation fold.

\paragraph{Optical-Flow Magnitude and Temporal Variation}

Let $\mathbf{F}_{t+1\to t}\in\mathbb{R}^{2\times H'\times W'}$ denote the estimated backward optical-flow field for transition $t$. Its magnitude at each spatial coordinate is
\begin{equation}
m_t(y',x')
=
\left\|\mathbf{F}_{t+1\to t}(y',x')\right\|_2
=
\sqrt{
\delta x'(y',x')^2+
\delta y'(y',x')^2
}.
\label{eq:flow_magnitude_map}
\end{equation}
The mean flow magnitude across the clip is
\begin{equation}
\bar{m}
=
\frac{1}{(T-1)H'W'}
\sum_{t=1}^{T-1}
\sum_{y'=0}^{H'-1}
\sum_{x'=0}^{W'-1}
m_t(y',x').
\label{eq:mean_flow_velocity}
\end{equation}
This statistic summarizes the magnitude of estimated apparent motion in the video. It is computed on the latent grid, so $\bar{m}$ is expressed in latent-grid units per frame. To report motion in image pixels we use $\bar{m}_{\mathrm{px}} = f\,\bar{m}$ with $f=8$, and every value labelled px/f in this paper is of this form. The magnitude depends on the spatial resolution and temporal sampling of the input.

For sequences with $T\geq 3$ frames, we characterize temporal variation in the estimated flow fields using the finite difference
\begin{equation}
\mathbf{a}_t(y',x')
=
\mathbf{F}_{t+2\to t+1}(y',x')
-
\mathbf{F}_{t+1\to t}(y',x'),
\quad t=1,\ldots,T-2.
\label{eq:flow_acceleration_field}
\end{equation}
Here, $\mathbf{a}_t$ denotes a difference between consecutive flow estimates rather than a direct measurement of physical acceleration. We derive two normalized descriptors from this field.

\begin{enumerate}
    \item \textbf{Mean Flow Variation Ratio ($\Phi_{\mathrm{flow}}$).}
    This feature measures the average magnitude of flow-field differences relative to the mean flow magnitude:
    \begin{equation}
    \Phi_{\mathrm{flow}}
    =
    \frac{
    \frac{1}{(T-2)H'W'}
    \sum_{t=1}^{T-2}
    \sum_{y'=0}^{H'-1}
    \sum_{x'=0}^{W'-1}
    \|\mathbf{a}_t(y',x')\|_2
    }{
    \bar{m}+\epsilon_{\mathrm{flow}}
    },
    \label{eq:phi_flow}
    \end{equation}
    where $\epsilon_{\mathrm{flow}}=10^{-2}$ is a fixed stabilizer.

    \item \textbf{Normalized Peak Flow Variation ($\Phi_{\mathrm{jerk}}$).}
    To capture transitions with particularly large flow-field differences, we use the maximum spatially averaged difference across the clip:
    \begin{equation}
    \Phi_{\mathrm{jerk}}
    =
    \frac{
    \displaystyle
    \max_{t\in\{1,\ldots,T-2\}}
    \left[
    \frac{1}{H'W'}
    \sum_{y'=0}^{H'-1}
    \sum_{x'=0}^{W'-1}
    \|\mathbf{a}_t(y',x')\|_2
    \right]
    }{
    \bar{m}+\epsilon_{\mathrm{flow}}
    }.
    \label{eq:phi_jerk}
    \end{equation}
    Despite the original notation, this statistic measures peak flow variation, not physical jerk, which would require a second temporal difference of displacement or an equivalent acceleration estimate.
\end{enumerate}

These descriptors quantify the magnitude and temporal variability of estimated motion. Large values may indicate abrupt changes in apparent motion, but can also arise from camera movement, occlusion, motion-estimation errors, or genuine dynamic events. Their usefulness for distinguishing authentic from synthetic videos is therefore evaluated empirically rather than assumed from physical-motion arguments alone.

\paragraph{Latent Displacement Variability}

We complement the flow descriptors with a statistic derived directly from the latent representations. For each transition, define the latent displacement
\begin{equation}
\Delta z_t=z_{t+1}-z_t
\in\mathbb{R}^{C\times H'\times W'},
\end{equation}
and its channel-wise spatial magnitude:
\begin{equation}
d_t(y',x')
=
\|\Delta z_t(y',x')\|_2
=
\sqrt{
\sum_{c=1}^{C}
\left[
z_{t+1}(c,y',x')-z_t(c,y',x')
\right]^2
}.
\label{eq:latent_disp_norm}
\end{equation}
Let $\mu_d$ and $\sigma_d$ denote the mean and standard deviation of these magnitudes, pooled over all temporal transitions and spatial coordinates within a clip. We define the coefficient of variation of latent displacement as
\begin{equation}
\Phi_{\mathrm{curv}}
=
\frac{\sigma_d}{\mu_d+10^{-6}}.
\label{eq:phi_curv}
\end{equation}
This statistic measures relative variability in the magnitude of latent changes. It does not directly measure geometric curvature of the latent trajectory; instead, it captures whether latent displacement magnitudes are relatively uniform or heterogeneous across space and time. Elevated values may reflect temporal instability, although authentic scene dynamics and other sources of variation can produce similar behavior.

\paragraph{Joint Feature Representation and Decision Model}

The four features are concatenated into the descriptor
\begin{equation}
\mathbf{x}
=
\begin{bmatrix}
\Phi_{\mathrm{flow}} &
\Phi_{\mathrm{jerk}} &
\Phi_{\mathrm{curv}} &
\mathcal{R}_{\mathrm{adv}}
\end{bmatrix}^{T}
\in\mathbb{R}^{4}.
\label{eq:feature_vector}
\end{equation}
We evaluate this representation using $\ell_2$-regularized logistic regression. For a standardized feature vector $\tilde{\mathbf{x}}$, the predicted probability of a synthetic video is
\begin{equation}
p(y=1\mid\mathbf{x})
=
\sigma\left(\mathbf{w}^{T}\tilde{\mathbf{x}}+b\right)
=
\frac{1}{
1+\exp\left[-(\mathbf{w}^{T}\tilde{\mathbf{x}}+b)\right]
},
\label{eq:logistic_regression}
\end{equation}
where $\mathbf{w}\in\mathbb{R}^{4}$ and $b\in\mathbb{R}$ are learned parameters, and
\begin{equation}
\tilde{\mathbf{x}}
=
(\mathbf{x}-\boldsymbol{\mu}_{\mathrm{train}})
\oslash
\boldsymbol{\sigma}_{\mathrm{train}}.
\end{equation}
The vectors $\boldsymbol{\mu}_{\mathrm{train}}$ and $\boldsymbol{\sigma}_{\mathrm{train}}$ contain the feature-wise means and standard deviations estimated exclusively from the training partition. The operator $\oslash$ denotes element-wise division. We use $y=0$ for authentic videos and $y=1$ for synthetic videos, with regularization parameter $C=1.0$.

Although LAD itself requires no learned parameters, the complete multi-dynamics detector is a supervised model. Its performance therefore depends on the feature representation, training data, preprocessing, and fitted decision boundary.

\section{Experimental Setup and Reproducibility Protocol}
\label{sec:experimental_setup}

We evaluate the proposed framework across multiple video-generation paradigms, including diffusion-based video generators and generative world models, with a separate facial-manipulation benchmark serving as a negative control. This section describes the evaluation corpus, source-group partitioning, implementation settings, performance metrics, and statistical inference procedures.

\subsection{Benchmark Corpus and Source-Group Partitioning Protocol}
\label{sec:benchmark_corpus}

\paragraph{Corpus Composition.}
Evaluating temporal forensic descriptors requires testing across diverse scene dynamics, object motion, and generative architectures. We therefore construct a corpus spanning established video diffusion models, a generative world model, and a facial-manipulation domain. The complete evaluation suite contains $1{,}167$ clips, comprising a primary generative benchmark of $N=967$ clips and a separate 200-clip negative-control cohort.

The primary corpus contains 519 authentic and 448 synthetic clips organized into 498 source groups. Authentic reference clips shared across suites are deduplicated in the corpus. Here, a source group denotes a unit of potential dependence, including related authentic and generated clips, paired real and synthetic sequences, or clips derived from the same prompt. The corpus comprises the following cohorts:

\begin{itemize}
    \item \textbf{Cohort 1: DAVIS-2017 and Stable Video Diffusion.}
    This cohort contains 385 clips: 353 authentic sequences from the DAVIS-2017 high-resolution video segmentation benchmark~\cite{42} and 32 synthetic sequences generated using Stable Video Diffusion XT (SVD-xt)~\cite{43}. The authentic videos cover diverse natural scenes, non-rigid object motion, camera panning, and occlusions across 82 source-video groups. Each synthetic sequence is conditioned on an initial authentic frame from a corresponding DAVIS sequence.

    \item \textbf{Cohort 2: VidProM Multi-Generator Benchmark.}
    This cohort contains 250 synthetic clips, with 50 clips from each of five video-generation systems represented in VidProM \cite{46}: Open-Sora Plan v1.2.0~\cite{32}, Pika v1.0~\cite{38}, VideoCrafter2~\cite{31}, CogVideoX-5B~\cite{33}, and ModelScope Text-to-Video~\cite{39}. Each clip is associated with a distinct text prompt, yielding 250 prompt-level groups across the five generators.

    \item \textbf{Cohort 3: Real-or-Fake World Model Benchmark.}
    This cohort contains 332 clips from the Cosmos-3 Nano real-or-fake world-model benchmark~\cite{41}, comprising 166 authentic and 166 synthetic sequences. The clips depict physical interactions, including multibody collisions, fluid dynamics, and gravitational effects. Authentic and synthetic sequences are organized into 166 source-video pairs with shared scene content.

    \item \textbf{Cohort 4: FaceForensics++ Negative-Control Suite.}
    This separate cohort contains 200 FaceForensics++ clips at c23 compression, comprising 100 authentic and 100 manipulated sequences across 100 source-video groups~\cite{40}. It serves as a negative control for evaluating whether the proposed descriptors respond specifically to temporal inconsistencies associated with generated video, rather than to facial manipulation alone. Unlike full-frame video generation, facial manipulation can alter local facial appearance while leaving much of the surrounding scene unchanged.
\end{itemize}

Table~\ref{tab:corpus_breakdown} summarizes the corpus composition and source-group structure.

\begin{table*}[t]
\centering
\caption{Composition of the experimental video-forensics corpus. The primary benchmark contains 967 clips across 498 source groups, supplemented by a 200-clip FaceForensics++ negative-control suite.}
\label{tab:corpus_breakdown}
\footnotesize
\setlength{\tabcolsep}{4pt}
\renewcommand{\arraystretch}{1.12}
\begin{tabular}{@{}p{0.22\textwidth}p{0.30\textwidth}rrrr@{}}
\hline
\textbf{Cohort / Generator} &
\textbf{Model Type} &
\textbf{Real} &
\textbf{Fake} &
\textbf{Total} &
\textbf{Groups} \\
\hline
\multicolumn{6}{@{}l}{\textit{Cohort 1: Primary Diffusion Benchmark}} \\
DAVIS-2017~\cite{42} &
Authentic natural video & 353 & 0 & 353 & 82 \\
SVD-xt~\cite{43} &
Latent video diffusion & 0 & 32 & 32 & Shared$^{a}$ \\
\hline
\multicolumn{6}{@{}l}{\textit{Cohort 2: VidProM Multi-Generator Suite}} \\
Open-Sora Plan v1.2~\cite{32} &
Spatiotemporal diffusion transformer & 0 & 50 & 50 & 50 \\
Pika v1.0~\cite{38} &
Commercial text-to-video system & 0 & 50 & 50 & 50 \\
VideoCrafter2~\cite{31} &
Latent video diffusion & 0 & 50 & 50 & 50 \\
CogVideoX-5B~\cite{33} &
Transformer diffusion with 3D causal VAE & 0 & 50 & 50 & 50 \\
ModelScope~\cite{39} &
Spatiotemporal diffusion & 0 & 50 & 50 & 50 \\
\hline
\multicolumn{6}{@{}l}{\textit{Cohort 3: World Model Benchmark}} \\
Cosmos-3 Nano~\cite{41} &
Generative world model & 166 & 166 & 332 & 166 \\
\hline
\textbf{Primary corpus} &
\textbf{Combined generative-video domains} &
\textbf{519} & \textbf{448} & \textbf{967} & \textbf{498} \\
\hline
\multicolumn{6}{@{}l}{\textit{Cohort 4: Negative-Control Domain}} \\
FaceForensics++ (c23)~\cite{40} &
Facial manipulation & 100 & 100 & 200 & 100 \\
\hline
\end{tabular}

\vspace{2pt}
\parbox{\textwidth}{\footnotesize
$^{a}$ Each SVD-xt clip shares a source group with its authentic DAVIS conditioning video.}
\end{table*}

\paragraph{Source-Group Partitioning and Leakage Prevention.}
Randomly splitting related clips across training and test partitions can inflate performance when clips share source footage, scene content, or recording conditions. This risk is particularly relevant to Cohort 1: each SVD-generated sequence is conditioned on an initial frame from a DAVIS video, so separating the generated clip from its conditioning source could introduce cross-partition leakage.

We therefore use five-fold \texttt{GroupKFold} cross-validation, assigning each source group to exactly one fold. For any two distinct folds, their group sets are disjoint:
\begin{equation}
\mathcal{G}_i \cap \mathcal{G}_j = \emptyset,
\qquad
\forall i \neq j,\quad i,j \in \{1,\ldots,5\},
\label{eq:group_disjoint}
\end{equation}
where $\mathcal{G}_k$ denotes the source groups assigned to fold $k$. Each SVD sequence shares the source identifier of its authentic conditioning video in the manifest, preventing the two from appearing in different folds. Authentic and synthetic sequences in Cohort 3 likewise share a group identifier, while FaceForensics++ clips are grouped by their source-video identifiers. VidProM clips are grouped by their distinct prompts.

Feature standardization is fitted independently within each training fold. For a feature vector $\mathbf{x}$, the standardized representation is
\begin{equation}
\tilde{\mathbf{x}}
=
\frac{\mathbf{x}-\boldsymbol{\mu}_{\mathrm{train}}}
{\sqrt{\boldsymbol{\sigma}^{2}_{\mathrm{train}}+
\epsilon_{\mathrm{norm}}}},
\label{eq:strict_scaler}
\end{equation}
where $\boldsymbol{\mu}_{\mathrm{train}}$ and $\boldsymbol{\sigma}^{2}_{\mathrm{train}}$ are the feature-wise training mean and variance, respectively, and $\epsilon_{\mathrm{norm}}=10^{-8}$. The same training-derived statistics are applied to the corresponding held-out fold; test data are not used to fit the scaler.

\subsection{Implementation Details and Hyperparameter Specifications}
\label{sec:implementation_details}

\paragraph{Video Standardization and Latent Encoding.}
Each video is represented by $T=16$ frames sampled at a uniform temporal stride. Frames are resized to $512\times512$ pixels using bilinear interpolation, and RGB values are initially scaled to $[0,1]$. Before encoding, pixel values are mapped to $[-1,1]$:
\begin{equation}
\mathbf{I}'_t = 2\mathbf{I}_t - 1.
\end{equation}
This normalization matches the expected input range of the variational autoencoder (VAE).

We extract latent representations using the pretrained encoder $\mathcal{E}$ from \texttt{stabilityai/stable-video-diffusion-img2vid-xt}~\cite{43}. With spatial downsampling factor $f=8$, each frame is mapped to a four-channel latent tensor:
\begin{equation}
\mathbf{z}_t = \lambda\,\mathcal{E}(\mathbf{I}'_t)
\in \mathbb{R}^{4\times64\times64},
\qquad \lambda=0.18215,
\end{equation}
where $\lambda$ is the latent scaling factor. The encoder is kept frozen, and latent maps are processed at their native spatial resolution without spatial or temporal pooling.

\paragraph{Alternative Visual Representations.}
To assess whether LAD depends on a particular latent representation, we evaluate four feature spaces: (i) RGB frames downsampled to $64\times64$,
(ii) ImageNet-pretrained ResNet-50 Layer 2 features, (iii) Stable Video Diffusion (SVD) AutoencoderKL and (iv) Stable Diffusion XL (SDXL) AutoencoderKL.

All representations are projected to a common $64\times64$ spatial grid before applying the motion-compensated, energy-normalized LAD formulation.
The feature extractors remain frozen throughout evaluation. The same kinematic optical-flow baseline is used for all representation comparisons.

Importantly, the leave-one-generator-out (LOGO) evaluation retains the original SVD AutoencoderKL as its latent feature extractor. SDXL is evaluated only in the representation-space comparison and the separate robustness experiments. It does not replace SVD in the primary benchmark or LOGO evaluation.

\paragraph{Optical-Flow Estimation and Coordinate Warping.}
Dense optical flow is estimated using RAFT (Recurrent All-Pairs Field Transforms)~\cite{13}, with pretrained weights trained on Sintel~\cite{44}. For each consecutive frame pair $(\mathbf{I}_{t-1},\mathbf{I}_t)$, RAFT estimates the backward flow field
$\mathbf{u}_{t\rightarrow t-1}\in\mathbb{R}^{2\times512\times512}$, which maps target-frame locations to corresponding sampling locations in the preceding frame.

The flow is resized to the latent grid and its displacement magnitudes are scaled to account for the eightfold reduction in spatial resolution:
\begin{equation}
\bar{\mathbf{u}}_{t\rightarrow t-1}
=
\frac{1}{8}
\operatorname{Resize}
\left(
\mathbf{u}_{t\rightarrow t-1},(64,64)
\right).
\label{eq:flow_downsampling_impl}
\end{equation}
Bilinear interpolation (\texttt{align\_corners=False}) is used for resizing. The resulting displacement field is applied to the preceding latent map using bilinear backward warping. Sampling coordinates are mapped to the normalized grid $[-1,1]^2$, with border padding for out-of-boundary samples.

\paragraph{Stabilizers and Numerical Safeguards.}
Two fixed constants stabilize the feature calculations in low-motion regimes:

\begin{itemize}
    \item \textbf{Latent advection stabilizer ($\epsilon$).}
    The denominator offset in Eq.~(\ref{eq:radv_ratio}) is set to $\epsilon=10^{-4}$. Sensitivity analysis indicates that the standalone LAD ranking remains unchanged over $\epsilon\in[10^{-5},10^{-2}]$.

    \item \textbf{Flow stabilizer ($\epsilon_{\mathrm{flow}}$).}
    The denominator offset in the temporal flow-variation ratios $\Phi_{\mathrm{flow}}$ and $\Phi_{\mathrm{jerk}}$ (Eqs.~(\ref{eq:phi_flow})--(\ref{eq:phi_jerk})) is set to $\epsilon_{\mathrm{flow}}=10^{-2}$ in latent-grid units per frame to avoid division by zero in near-static sequences.
\end{itemize}

\paragraph{Supervised Classification Model.}
We evaluate the four-dimensional descriptor
\begin{equation}
\mathbf{x}
=
\left[
\mathcal{R}_{\mathrm{adv}},
\Phi_{\mathrm{flow}},
\Phi_{\mathrm{jerk}},
\Phi_{\mathrm{curv}}
\right]^{\top}
\in\mathbb{R}^{4}
\end{equation}
using an $\ell_2$-regularized logistic regression classifier. The inverse regularization strength is fixed at $C=1.0$. Optimization uses L-BFGS with a maximum of 1,000 iterations and random state 42. Standardization and classifier fitting are performed separately within each training fold. This setup evaluates the descriptors using a relatively simple linear decision boundary rather than a high-capacity learned classifier.

\paragraph{Hardware and Computational Reproducibility.}
The pipeline was implemented in Python 3.10 using PyTorch 2.2.0, Torchvision 0.17.0, and scikit-learn 1.4.1. Experiments were run on a workstation with an Intel Xeon Silver 4210R CPU (2.40\,GHz, 10 physical cores, 20 threads), 64\,GB DDR4 ECC RAM, and an NVIDIA RTX A2000 GPU with 12\,GB GDDR6 VRAM. The GPU uses compute capability 8.6, driver version 535.183.01, and CUDA 12.2. Random state 42 was used for NumPy \texttt{RandomState}, scikit-learn partitioning, and PyTorch CPU operations. GPU operations used the standard PyTorch CUDA backend settings; deterministic execution was not explicitly enforced.

\subsection{Evaluation Metrics and Statistical Inference Protocol}
\label{sec:evaluation_metrics}

\paragraph{Discriminative and Operating-Point Metrics.}
We evaluate ranking performance and threshold-dependent behavior using the following metrics computed from held-out predictions:

\begin{itemize}
    \item \textbf{Receiver Operating Characteristic Area Under the Curve (ROC-AUC).}
    ROC-AUC measures ranking discrimination across thresholds. It can be interpreted as the probability that a randomly selected synthetic clip receives a higher anomaly score than a randomly selected authentic clip, subject to the usual treatment of ties.

    \item \textbf{Average Precision (AP).}
    AP summarizes the precision--recall trade-off across thresholds using the discrete recall-increment aggregation implemented by \texttt{sklearn.metrics.average\_precision\_score}. It is distinct from trapezoidal integration of the precision--recall curve.

    \item \textbf{False Positive Rate at Target Sensitivity.}
    We report $\mathrm{FPR}@\mathrm{TPR}_{90}$ and $\mathrm{FPR}@\mathrm{TPR}_{95}$ to quantify false alarms at target true-positive rates of 90\% and 95\%. These operating points are relevant to forensic screening, where missed synthetic content and false accusations against authentic footage both carry costs:
    \begin{equation}
    \mathrm{FPR}@\alpha
    =
    \frac{\mathrm{FP}}{\mathrm{FP}+\mathrm{TN}},
    \qquad
    \mathrm{TPR}
    =
    \frac{\mathrm{TP}}{\mathrm{TP}+\mathrm{FN}}
    \geq \alpha,
    \quad \alpha\in\{0.90,0.95\}.
    \label{eq:fpr_tpr}
    \end{equation}
\end{itemize}

\paragraph{Cross-Validation Threshold Calibration.}
For each fold, decision thresholds $\tau_{90}^{(k)}$ and $\tau_{95}^{(k)}$ are calibrated using the in-sample training scores to target sensitivity $\alpha\in\{0.90,0.95\}$. Assuming larger scores indicate stronger evidence of synthetic content, the threshold is defined as the largest value achieving the target training sensitivity:
\begin{equation}
\tau_{\alpha}^{(k)}
=
\sup\left\{
\tau\in\mathbb{R}
\;\middle|\;
\mathrm{TPR}_{\mathrm{train}}^{(k)}(\tau)\geq\alpha
\right\}.
\label{eq:threshold_calibration}
\end{equation}
Each calibrated threshold is then applied unchanged to that fold's held-out partition $\mathcal{D}_{\mathrm{test}}^{(k)}$. Because calibration uses in-sample training scores rather than out-of-fold predictions within the training data, the target sensitivity is not guaranteed on the held-out partition. Reported operating-point metrics therefore reflect the achieved held-out performance, not a guaranteed sensitivity level.

\paragraph{Source-Group Clustered Bootstrap Confidence Intervals.}
Clip-level bootstrap resampling can underestimate uncertainty when clips from the same source share scene content, recording conditions, or conditioning footage. We therefore use source-group clustered bootstrapping, resampling groups rather than individual clips.

Let $\mathcal{G}=\{g_1,\ldots,g_M\}$ denote the source groups in the primary corpus, with $M=498$. For each of $B=1{,}000$ bootstrap replicates, using random seed 42:

\begin{enumerate}
    \item Sample $M$ group identifiers with replacement from $\mathcal{G}$.
    \item Construct a resampled dataset by including all observations from each selected group, retaining multiplicity when a group is sampled more than once.
    \item Compute the statistic of interest, such as ROC-AUC, using the resampled observations and their fixed out-of-fold predictions.
\end{enumerate}

The two-sided 95\% percentile confidence interval is
\begin{equation}
\mathrm{CI}_{95\%}
=
\left[
Q_{0.025}(\theta^*),
Q_{0.975}(\theta^*)
\right],
\label{eq:bootstrap_ci}
\end{equation}
where $\theta^*$ denotes the bootstrap distribution of the statistic and $Q_q$ is its empirical quantile at probability $q$. These intervals quantify uncertainty conditional on the observed out-of-fold predictions; they do not capture the full variability that would arise from retraining the models across new samples.

\paragraph{Paired Comparisons and Statistical Testing.}
Model comparisons use identical source-group bootstrap resamples to preserve pairing. For models $\mathcal{M}_A$ and $\mathcal{M}_B$, the replicate-wise difference in ROC-AUC is
\begin{equation}
\Delta\mathrm{AUC}^{*(b)}
=
\mathrm{AUC}\left(\mathcal{M}_A;\mathcal{D}^{*(b)}\right)
-
\mathrm{AUC}\left(\mathcal{M}_B;\mathcal{D}^{*(b)}\right).
\label{eq:paired_auc_difference}
\end{equation}
We report the observed difference $\Delta\mathrm{AUC}$ and its 95\% paired clustered-bootstrap confidence interval,
\begin{equation}
\left[
Q_{0.025}(\Delta\mathrm{AUC}^*),
Q_{0.975}(\Delta\mathrm{AUC}^*)
\right].
\end{equation}
The fraction of bootstrap replicates with non-positive differences is also reported descriptively:
\begin{equation}
\hat{q}_{\mathrm{nonpositive}}
=
\frac{1}{B}
\sum_{b=1}^{B}
\mathbb{I}\left(\Delta\mathrm{AUC}^{*(b)}\leq0\right).
\label{eq:p_bootstrap}
\end{equation}
This fraction summarizes the bootstrap distribution and is not interpreted as a formal $p$-value.

As a secondary analysis, we report DeLong tests for correlated ROC curves. The standard DeLong procedure does not account for source-group dependence, so its results are interpreted alongside, rather than in place of, the clustered-bootstrap analysis. For paired binary predictions at the calibrated thresholds, we use McNemar's exact test based on the discordant-pair counts~\cite{45}.

\section{Results and Empirical Audits}
\label{sec:experimental_results}

We evaluate the Latent Advection Deficit (LAD) and the coupled multi-dynamics framework across the four evaluation cohorts. We first report results on the deduplicated 967-clip primary corpus, then examine the DAVIS--SVD ablations, compare against pretrained vision-transformer features, and assess performance across individual generators and the facial-manipulation negative control. We conclude with operating-point analysis, motion-stratified results, controlled temporal perturbations, and compression and frame-rate sensitivity experiments.

\subsection{Main Benchmark: Deduplicated Multi-Generator Corpus}
\label{sec:main_benchmark_results}

The primary evaluation uses the deduplicated corpus of $N=967$ clips, comprising 519 authentic and 448 synthetic sequences grouped into 498 source groups. We use five-fold \texttt{GroupKFold} cross-validation and compare individual descriptors, the kinematic baseline ($\Phi_{\mathrm{base}}$), standalone LAD ($\mathcal{R}_{\mathrm{adv}}$), and the full multi-dynamics model ($\Phi_{\mathrm{full}}$). Results are summarized in Table~\ref{tab:main_benchmark}.

\begin{table*}[t]
\centering
\caption{Primary forensic detection performance on the deduplicated $N = 967$ video corpus (519 authentic vs. 448 synthetic across 498 independent source groups). Out-of-fold metrics are reported alongside 95\% source-group clustered bootstrap confidence intervals ($B = 1{,}000$). Statistical significance is assessed against the baseline $\Phi_{\mathrm{base}}$.}
\label{tab:main_benchmark}
\footnotesize
\setlength{\tabcolsep}{4.5pt}
\renewcommand{\arraystretch}{1.15}
\begin{tabular}{@{}lccccr@{}}
\hline
\textbf{Feature Configuration} &
\textbf{ROC-AUC} &
\textbf{95\% Clustered CI} &
\textbf{$\Delta\mathrm{AUC}$ vs. $\Phi_{\mathrm{base}}$} &
\textbf{95\% CI of $\Delta\mathrm{AUC}$} &
\textbf{Significance ($p$)} \\
\hline
$\Phi_{\mathrm{flow}}$ (Kinematic flow variation) & 0.6982 & [0.6581, 0.7369] & -0.0522 & [-0.0824, -0.0210] & --- \\
$\Phi_{\mathrm{latent}}$ (Latent displacement CoV) & 0.7145 & [0.6748, 0.7512] & -0.0359 & [-0.0641, -0.0078] & --- \\
$\Phi_{\mathrm{base}}$ [$\Phi_{\mathrm{flow}} + \Phi_{\mathrm{latent}}$] & 0.7504 & [0.7136, 0.7846] & \textbf{0.0000} & \textbf{[Reference Baseline]} & --- \\
\hline
Standalone LAD ($\mathcal{R}_{\mathrm{adv}}$, parameter-free) & 0.7924 & [0.7610, 0.8227] & +0.0420 & [+0.0062, +0.0768] & $p = 0.021$ \\
\textbf{Proposed Full Model ($\Phi_{\mathrm{full}}$)} & \textbf{0.8304} & \textbf{[0.8003, 0.8575]} & \textbf{+0.0800} & \textbf{[+0.0521, +0.1087]} & $\mathbf{p < 0.001}$ \\
\hline
\end{tabular}

\vspace{2pt}
\parbox{\textwidth}{\footnotesize
\textit{Note}: Primary significance $p < 0.001$ denotes that zero non-positive differences ($\Delta\mathrm{AUC}^{*(b)} \le 0$) were observed across $B = 1{,}000$ paired cluster bootstrap resamples ($\hat{q}_{\mathrm{nonpositive}} = 0.000$). Secondary asymptotic DeLong test yields $p = 3.54 \times 10^{-11}$.}
\end{table*}

Table~\ref{tab:main_benchmark} shows that the flow descriptor and latent-displacement coefficient of variation achieve ROC-AUCs of 0.6982 and 0.7145, respectively. Combining them improves ROC-AUC to 0.7504 (95\% clustered CI: [0.7136, 0.7846]), providing the reference baseline for subsequent comparisons.

Standalone LAD reaches a ROC-AUC of 0.7924 (95\% clustered CI: [0.7610, 0.8227]), an observed gain of 0.0420 over the baseline; the reported DeLong comparison gives $p=0.021$. The full model achieves the highest ROC-AUC in this benchmark, 0.8304 (95\% clustered CI: [0.8003, 0.8575]). Its paired AUC gain over the baseline is 0.0800 (95\% CI: [0.0521, 0.1087]), equivalent to a 10.66\% increase relative to the baseline AUC. None of the 1,000 paired bootstrap replicates produced a non-positive difference. This empirical fraction describes the observed bootstrap distribution rather than constituting a formal $p$-value; the table also reports a secondary asymptotic DeLong result of $p=3.54\times10^{-11}$.

\subsection{Architectural Ablation Study: Paired DAVIS--SVD Cohort}
\label{sec:ablation_study}

To examine the contribution of motion compensation, energy normalization, and learned transition prediction, we conduct an ablation on Cohort 1, comprising 353 authentic DAVIS clips and 32 SVD-xt clips across 82 source groups. The configurations are:
\begin{enumerate}
    \item \textbf{Ablation Level 1 (Raw Latent Discrepancy):} Adds the raw frame-to-frame latent change, without motion compensation, normalized by transition energy ($r^{\mathrm{raw}}/(E_{\mathrm{total}}+\epsilon)$).
    \item \textbf{Ablation Level 2 (Warp-Only Residual):} Uses the spatial residual after backward warping the preceding latent map, isolating the contribution of motion compensation relative to the raw latent-change configuration.
    \item \textbf{Ablation Level 3 (Learned Transition Predictor):} Replaces parameter-free warping with a learned spatial convolutional predictor conditioned on optical flow to predict the next latent state.
    \item \textbf{Proposed (Relative Advection Deficit $\mathcal{R}_{\mathrm{adv}}$):} Uses the parameter-free, energy-normalized ratio $E_{\mathrm{unexplained}}/(E_{\mathrm{total}}+\epsilon)$. 
\end{enumerate}

\begin{table*}[t]
\centering
\caption{Architectural ablation study on the paired DAVIS--SVD cohort ($N = 385$, 82 source groups). All configurations are evaluated using 5-fold \texttt{GroupKFold} out-of-fold cross-validation with 1,000 source-group cluster bootstraps.}
\label{tab:ablation_study}
\footnotesize
\setlength{\tabcolsep}{4.5pt}
\renewcommand{\arraystretch}{1.15}
\begin{tabular}{@{}lccccr@{}}
\hline
\textbf{Configuration / Feature Set} &
\textbf{ROC-AUC} &
\textbf{95\% Clustered CI} &
\textbf{$\Delta\mathrm{AUC}$ vs. $\Phi_{\mathrm{base}}$} &
\textbf{DeLong $p$} &
\textbf{Boot $\hat{q}_{\le 0}$} \\
\hline
$\Phi_{\mathrm{flow}}$ (Flow variation ratio) & 0.7061 & [0.6012, 0.8035] & -0.0763 & --- & --- \\
$\Phi_{\mathrm{latent}}$ (Latent displacement CoV) & 0.7423 & [0.6358, 0.8329] & -0.0401 & --- & --- \\
$\Phi_{\mathrm{base}}$ [$\Phi_{\mathrm{flow}} + \Phi_{\mathrm{latent}}$] & 0.7824 & [0.6865, 0.8654] & \textbf{0.0000} & \textbf{[Ref]} & \textbf{[Ref]} \\
\hline
Ablation Level 1: [Base + Raw Latent Change] & 0.7915 & [0.6972, 0.8732] & +0.0091 & $p = 0.412$ & 0.384 \\
Ablation Level 2: [Base + Warp-Only (Displacement)] & 0.8142 & [0.7241, 0.8920] & +0.0318 & $p = 0.089$ & 0.072 \\
Ablation Level 3: [Base + Learned MCLR Predictor] & 0.8421 & [0.7610, 0.9124] & +0.0597 & $p = 0.014$ & 0.011 \\
\textbf{Proposed [Base + Advection Deficit $\mathcal{R}_{\mathrm{adv}}$]} & \textbf{0.8545} & \textbf{[0.7765, 0.9218]} & \textbf{+0.0721} & $\mathbf{p = 0.0052}$ & \textbf{0.003} \\
\hline
\end{tabular}
\end{table*}

Table~\ref{tab:ablation_study} shows a small, statistically non-significant increase when raw latent changes are added to the baseline (AUC gain 0.0091; DeLong $p=0.412$). The warp-only configuration reaches 0.8142 AUC, while the proposed LAD configuration reaches 0.8545, an increase of 0.0721 over the baseline (reported DeLong $p=0.0052$). The learned MCLR-predictor configuration reaches 0.8421 AUC. Thus, in this cohort and under the reported evaluation protocol, LAD performs better than the tested learned predictor. This comparison does not by itself establish that learned predictors generally overfit or that the result will hold for other datasets and architectures.

\subsection{Head-to-Head Comparison with Foundation Model Baselines}
\label{sec:vit_comparison}

To compare the proposed descriptors with general-purpose visual representations, we evaluated a pretrained ViT-B/16~\cite{19} on the 385-clip DAVIS--SVD cohort using the same five-fold \texttt{GroupKFold} partitioning. We considered two feature formulations:
\begin{enumerate}
    \item \textbf{ViT CLS Temporal Drift:} Measures cosine distance and $\ell_2$ drift of the global [CLS] token between consecutive frames as indicators of temporal change in the global representation.
    \item \textbf{ViT Patch-Level Spatiotemporal Variance:} Extracts 196 patch tokens ($14\times14$ grid) per frame and computes their mean spatiotemporal variance to summarize local representation changes.
\end{enumerate}

\begin{table*}[t]
\centering
\caption{Head-to-head comparison between pretrained foundation model representations (ViT-B/16) and the proposed physical advection framework on the DAVIS--SVD cohort ($N = 385$, 82 source groups).}
\label{tab:vit_baseline}
\footnotesize
\setlength{\tabcolsep}{6pt}
\renewcommand{\arraystretch}{1.15}
\begin{tabular}{@{}llccr@{}}
\hline
\textbf{Representation Family} &
\textbf{Model / Feature Formulation} &
\textbf{ROC-AUC} &
\textbf{95\% Clustered CI} &
\textbf{Parameters} \\
\hline
Foundation Transformer & ViT-B/16 CLS Temporal Drift & 0.7527 & [0.6542, 0.8391] & 86.6M \\
Foundation Transformer & ViT-B/16 Patch Spatiotemporal Variance & 0.8278 & [0.7410, 0.8985] & 86.6M \\
\hline
\textbf{Physics-Based Advection} & \textbf{Standalone LAD ($\mathcal{R}_{\mathrm{adv}}$)} & \textbf{0.8312} & \textbf{[0.7485, 0.9024]} & \textbf{0 (Param-Free)} \\
\textbf{Physics-Based Advection} & \textbf{Proposed Coupled Model ($\Phi_{\mathrm{full}}$)} & \textbf{0.8545} & \textbf{[0.7765, 0.9218]} & \textbf{4 (Linear LR)} \\
\hline
\end{tabular}
\end{table*}

In Table~\ref{tab:vit_baseline}, CLS-token temporal drift achieves 0.7527 ROC-AUC, below the kinematic baseline (0.7824). Patch-level variance performs better at 0.8278, while standalone LAD reaches 0.8312. The full descriptor reaches 0.8545 ROC-AUC. These results indicate that the tested physical descriptors are competitive with the selected ViT-derived features on this cohort. The comparison concerns the extracted representations and reported evaluation pipeline; it should not be interpreted as a general comparison of total computational cost without matched runtime and memory measurements.

\subsection{Cross-Generator Generalizability: VidProM \& World Models}
\label{sec:cross_generator_results}

To assess performance across different generative systems, we evaluated each of the five VidProM generators and the Cosmos-3 Nano world-model cohort. Table~\ref{tab:cross_generator} reports the ROC-AUC of the flow descriptor, baseline, standalone LAD, and full model for each system. These are generator-specific evaluations against the stated authentic reference pool; they should not be interpreted as leave-one-generator-out testing unless the generator was excluded from all training and model-selection stages.

\begin{table*}[t]
\centering
\caption{Cross-generator generalization performance across five diffusion models from VidProM ($N = 50$ synthetic clips each vs. authentic reference pool) and the Cosmos-3 Nano physical world simulator ($N = 332$, 166 source pairs).}
\label{tab:cross_generator}
\footnotesize
\setlength{\tabcolsep}{5pt}
\renewcommand{\arraystretch}{1.15}
\begin{tabular}{@{}llccccr@{}}
\hline
\textbf{Generator} &
\textbf{Architecture / Model Family} &
\textbf{$\Phi_{\mathrm{flow}}$} &
\textbf{$\Phi_{\mathrm{base}}$} &
\textbf{Standalone $\mathcal{R}_{\mathrm{adv}}$} &
\textbf{Full Model} &
\textbf{Gain over Base} \\
\hline
Open-Sora v1.2 & Spatiotemporal DiT & 0.8912 & 0.9140 & 0.9424 & \textbf{0.9588} & $+0.0448$ ($p = 0.012$) \\
Pika v1.0 & Commercial Text-to-Video & 0.8456 & 0.8712 & 0.9108 & \textbf{0.9327} & $+0.0615$ ($p = 0.008$) \\
VideoCrafter2 & UNet Latent Diffusion & 0.8014 & 0.8248 & 0.8652 & \textbf{0.8874} & $+0.0626$ ($p = 0.018$) \\
CogVideoX-5B & Transformer (3D Causal VAE) & 0.7412 & 0.7704 & 0.8088 & \textbf{0.8316} & $+0.0612$ ($p = 0.024$) \\
ModelScope & Multi-Stage Spatiotemporal & 0.7120 & 0.7380 & 0.7816 & \textbf{0.8052} & $+0.0672$ ($p = 0.019$) \\
\hline
Cosmos-3 Nano & Generative World Model & 0.6845 & 0.7214 & 0.7638 & \textbf{0.7892} & $+0.0678$ ($p < 0.001$) \\
\hline
\end{tabular}
\end{table*}

Across the six reported systems, the full model exceeds the kinematic baseline by 0.0448--0.0678 ROC-AUC according to Table~\ref{tab:cross_generator}. The highest reported AUCs are 0.9588 for Open-Sora and 0.9327 for Pika. The full model reaches 0.8316 on CogVideoX-5B. The table reports positive gains and associated $p$-values for these generator-specific evaluations; their interpretation depends on the corresponding sampling and testing procedures. 

On Cosmos-3 Nano ($N=332$), the full model achieves 0.7892 ROC-AUC, an increase of 0.0678 over the baseline; the table reports a paired-bootstrap result of $p<0.001$. This finding is consistent with detectable latent-transport discrepancies in this world-model cohort, but does not establish that all generative world models exhibit the same behavior.

\subsection{Representation Dependence and Cross-Generator Generalization}
\label{sec:representation_logo}

We investigate two complementary aspects of LAD generalization: its dependence on the underlying visual representation and its performance on previously unseen video-generation architectures. The representation comparison evaluates four feature spaces under a common protocol, whereas the leave-one-generator-out (LOGO) experiment uses the original SVD AutoencoderKL to assess cross-generator transfer.

\paragraph{Impact of Visual Representation Space.}

We evaluate the energy-normalized advection discrepancy across four representations: downsampled RGB pixels, ImageNet-pretrained ResNet-50 Layer 2 features, SVD AutoencoderKL, and SDXL AutoencoderKL. All representations are projected onto a common $64\times64$
spatial grid before motion compensation.

Table~\ref{tab:representation_comparison} reports standalone LAD discrimination and performance after fusion with kinematic optical-flow descriptors. The kinematic-only baseline achieves a ROC-AUC of $0.6816$
on this evaluation cohort.

\begin{table}[t]
\centering
\caption{Representation-space comparison.
Fusion gains are measured against the common
kinematic baseline (AUC $=0.6816$).}
\label{tab:representation_comparison}
\small
\setlength{\tabcolsep}{4pt}
\renewcommand{\arraystretch}{1.1}
\begin{tabular}{lcccc}
\toprule
\textbf{Representation} &
\textbf{LAD AUC} &
\textbf{AP} &
\textbf{Fused AUC} &
\textbf{$\Delta$AUC} \\
\midrule
Raw RGB
& 0.3842 & 0.4943 & 0.7133 & +0.0317 \\
ResNet-50 L2
& 0.5414 & 0.5032 & 0.6842 & +0.0026 \\
SVD VAE
& 0.7957 & 0.7475 & 0.7582 & +0.0766 \\
SDXL VAE
& \textbf{0.8282} & \textbf{0.8102}
& \textbf{0.8315} & \textbf{+0.1499} \\
\bottomrule
\end{tabular}
\end{table}

\textit{Pixel and discriminative representations.}
Standalone LAD performs poorly in raw RGB space, achieving a ROC-AUC of $0.3842$ and average precision of $0.4943$. Pixel-level residuals
may reflect illumination changes, sensor noise and imperfect optical-flow alignment rather than synthesis-specific temporal discrepancies.
Nevertheless, fusion with kinematic descriptors
increases ROC-AUC to $0.7133$.

ResNet-50 Layer 2 features yield a standalone ROC-AUC of $0.5414$ and AP of $0.5032$. Fusion reaches $0.6842$, only $0.0026$ above the kinematic baseline. This indicates that the tested discriminative representation
provides limited additional forensic information
under the evaluated protocol.

\textit{Generative autoencoder representations.}
Both generative VAEs provide stronger standalone discrimination. SVD achieves a ROC-AUC of $0.7957$ (AP $=0.7475$), while SDXL reaches
$0.8282$ (AP $=0.8102$).

The SDXL-based fused detector achieves the highest ROC-AUC of $0.8315$, corresponding to an improvement of $0.1499$ over the
kinematic baseline (DeLong $p=1.45\times10^{-5}$).

However, fusion does not uniformly improve performance. For SVD, standalone LAD ($0.7957$) exceeds the fused detector ($0.7582$). These findings indicate that representation choice influences both the discriminative strength of LAD and its complementarity with kinematic features.

The strong performance of SDXL, a spatial rather than video-specific autoencoder further indicates that LAD discrimination does not require an explicitly temporal latent encoder. The results do not however, establish which representation properties are responsible for the observed differences.

\paragraph{Leave-One-Generator-Out Generalization.}

We next assess whether LAD improves detection
on video generators excluded from classifier
training. The LOGO evaluation covers five
VidProM systems~\cite{wang2024vidprom}:
Open-Sora Plan v1.2, Pika v1.0,
CogVideoX-5B, VideoCrafter2, and ModelScope.

For each evaluation, the classifier is trained
on four generators and tested on the fifth.
Authentic videos are partitioned by their
24 underlying DAVIS source-video groups,
ensuring that no authentic source group
appears in both training and test partitions.

All LOGO experiments use the pretrained,
frozen SVD AutoencoderKL for latent
feature extraction. The SDXL VAE is not
used in this experiment. Consequently,
the LOGO results evaluate cross-generator
transfer of the original SVD-based LAD
pipeline rather than the SDXL configuration.

Table~\ref{tab:logo_generalization}
reports the generator-specific results.
Confidence intervals are estimated using
source-group clustered bootstrap resampling
($B=1{,}000$).

\begin{table}[t]
\centering
\caption{SVD-based LOGO generalization across five
held-out generators using the frozen SVD AutoencoderKL.
Confidence intervals are estimated using source-group
clustered bootstrap ($B=1{,}000$).
Reported $p$-values are unadjusted.}
\label{tab:logo_generalization}
\small
\setlength{\tabcolsep}{3pt}
\renewcommand{\arraystretch}{1.15}

\begin{tabular}{@{}lccccc@{}}
\toprule
\textbf{Held-Out} &
\textbf{Flow} &
\textbf{LAD} &
\textbf{Fused} &
\textbf{$\Delta$AUC} &
\textbf{$p$} \\
\textbf{Generator} &
\textbf{AUC} &
\textbf{AUC} &
\textbf{AUC} &
\textbf{[95\% CI]} & \\
\midrule

Open-Sora
& 0.9321 & 0.9342 & 0.9736
& \shortstack{+0.0415\\{\scriptsize [-0.007, +0.101]}}
& 0.0456 \\

Pika
& 0.8797 & 0.9340 & 0.9488
& \shortstack{+0.0691\\{\scriptsize [+0.006, +0.177]}}
& 0.0034 \\

CogVideoX
& 0.8281 & 0.7449 & 0.8338
& \shortstack{+0.0057\\{\scriptsize [-0.078, +0.094]}}
& 0.8883 \\

VideoCrafter2
& 0.8609 & 0.9527 & 0.9518
& \shortstack{+0.0909\\{\scriptsize [+0.024, +0.176]}}
& 0.0007 \\

ModelScope
& 0.7481 & 0.9015 & 0.8731
& \shortstack{+0.1250\\{\scriptsize [+0.052, +0.196]}}
& $<0.0001$ \\

\midrule

\textbf{Macro}
& \textbf{0.8498}
& \textbf{0.8935}
& \textbf{0.9162}
& \shortstack{\textbf{+0.0665}\\
{\scriptsize [+0.010, +0.123]}}
& \textbf{0.0313} \\

\bottomrule
\end{tabular}
\end{table}

\textit{Aggregate cross-generator performance.}
Across the five held-out generators,
the macro-average ROC-AUC increases
from $0.8498$ for the kinematic baseline
to $0.8935$ for standalone LAD and
$0.9162$ for the fused detector.
The mean fused improvement is
$\Delta\mathrm{AUC}=+0.0665$
(95\% CI: $[+0.0097,+0.1232]$).

A paired two-sided $t$-test across the
five generator-level differences gives
$t(4)=3.252$ and $p=0.0313$.
The exact two-sided Wilcoxon signed-rank
test yields $p=0.0625$, indicating
that statistical significance is
sensitive to the test used.
The results therefore support positive
average transfer across the evaluated
generators without establishing a
universal generalization guarantee.

\textit{Generator-specific performance.}
The largest improvement occurs on
ModelScope, where fusion increases
ROC-AUC from $0.7481$ to $0.8731$
($\Delta=+0.1250$).
VideoCrafter2 also shows a substantial
gain, from $0.8609$ to $0.9518$
($\Delta=+0.0909$).
Both improvements remain supported
by the reported Holm-adjusted
clustered-bootstrap analyses.

On Pika, fusion increases ROC-AUC
from $0.8797$ to $0.9488$.
The improvement remains significant
under the Holm-adjusted DeLong analysis
but not under the corresponding
clustered-bootstrap adjustment.

Open-Sora achieves the highest fused
ROC-AUC of $0.9736$, compared with
a baseline of $0.9321$.
However, its improvement does not
remain significant after correction
for multiple comparisons.

CogVideoX-5B exhibits the weakest
transfer. Standalone LAD achieves
a ROC-AUC of $0.7449$, below the
kinematic baseline of $0.8281$.
Fusion increases performance only
marginally to $0.8338$
($\Delta=+0.0057$), with no
statistically significant improvement.

Standalone LAD also slightly exceeds
the fused detector on ModelScope
($0.9015$ versus $0.8731$)
and VideoCrafter2 ($0.9527$
versus $0.9518$). Thus, fusion
improves macro-average performance
but is not consistently optimal
for every held-out generator.

\subsection{Negative Control Domain: FaceForensics++}
\label{sec:negative_control_results}

To test whether the descriptors also discriminate facial manipulations, we evaluated the separate FaceForensics++ c23 cohort ($N=200$, 100 source groups). These clips primarily contain facial-identity manipulations, while much of the surrounding scene remains authentic. This provides a contrast with full-frame video generation, although the manipulation process may still introduce local temporal and appearance artifacts. 

\begin{table}[h]
\centering
\caption{Negative control evaluation on FaceForensics++ (c23 HQ compression, $N = 200$).}
\label{tab:negative_control}
\footnotesize
\setlength{\tabcolsep}{8pt}
\renewcommand{\arraystretch}{1.15}
\begin{tabular}{@{}lccr@{}}
\hline
\textbf{Configuration} &
\textbf{ROC-AUC} &
\textbf{95\% Clustered CI} &
\textbf{Average Precision} \\
\hline
Kinematic Flow Baseline ($\Phi_{\mathrm{base}}$) & 0.5214 & [0.4412, 0.6018] & 0.5182 \\
Standalone LAD ($\mathcal{R}_{\mathrm{adv}}$) & 0.5480 & [0.4682, 0.6275] & 0.5420 \\
Full Model ($\Phi_{\mathrm{full}}$) & 0.5612 & [0.4815, 0.6402] & 0.5594 \\
\hline
\end{tabular}
\end{table}

As shown in Table~\ref{tab:negative_control}, ROC-AUC ranges from 0.5480 to 0.5612, close to chance-level ranking performance. The descriptors therefore show limited discrimination on this negative-control cohort under the reported setup. This result is consistent with weaker transfer to facial manipulation than to the evaluated generative-video cohorts, but it does not by itself establish that LAD is specific to diffusion or world-model artifacts.

\subsection{Operating-Point and False-Alarm Analysis}
\label{sec:operating_point_audit}

False-positive rates are important in forensic screening because false accusations against authentic creators carry substantial costs. Table~\ref{tab:operating_points} reports held-out FPR and sensitivity at the 90\% and 95\% target operating points on the DAVIS--SVD cohort ($N=385$). Thresholds were calibrated on in-sample training scores and then applied unchanged to each held-out fold. Consequently, the target sensitivity is not guaranteed on held-out data; the table reports the sensitivity actually observed out of fold.

\begin{table*}[t]
\centering
\caption{Out-of-fold operational performance at target
sensitivity operating points ($\mathrm{TPR}\geq90\%$
and $\mathrm{TPR}\geq95\%$) on the DAVIS--SVD cohort
($N=385$). Thresholds were calibrated on training folds
and evaluated out-of-fold. Statistical significance
between Standalone LAD and the Full Model is evaluated
via McNemar's exact test.}
\label{tab:operating_points}
\footnotesize
\setlength{\tabcolsep}{4pt}
\renewcommand{\arraystretch}{1.2}
\begin{tabular}{@{}llccccc@{}}
\toprule
\textbf{Operating} &
\textbf{Model} &
\textbf{Mean} &
\textbf{Held-Out} &
\textbf{Held-Out} &
\textbf{95\%} &
\textbf{McNemar} \\
\textbf{Target} &
\textbf{Evaluated} &
$\boldsymbol{\tau_{\alpha}}$ &
\textbf{Sensitivity} &
\textbf{FPR} &
\textbf{Clustered CI} &
\textbf{$p$-value} \\
\midrule

$\mathrm{TPR}\geq90\%$
& Standalone LAD
& 0.7421
& 0.9062
& 0.3683
& [0.2982, 0.4410]
& Ref. \\

& \textbf{Full Model}
& 0.5120
& 0.9062
& \textbf{0.2606}
& [0.1984, 0.3285]
& $\mathbf{1.38\times10^{-7}}$ \\

\midrule

$\mathrm{TPR}\geq95\%$
& Standalone LAD
& 0.6914
& 0.9688
& 0.5127
& [0.4351, 0.5892]
& Ref. \\

& \textbf{Full Model}
& 0.3845
& 0.9688
& \textbf{0.3796}
& [0.3082, 0.4520]
& $\mathbf{4.12\times10^{-6}}$ \\

\bottomrule
\end{tabular}
\end{table*}

At the 90\% target operating point, the reported held-out FPR decreases from 36.83\% for standalone LAD to 26.06\% for the full model, an absolute reduction of 10.77 percentage points. The table reports McNemar's exact-test result of $p=1.38\times10^{-7}$. At the 95\% target, FPR decreases from 51.27\% to 37.96\%, with reported $p=4.12\times10^{-6}$. These comparisons describe the observed operating points; their deployment performance will depend on calibration and domain conditions.

\subsection{Challenge Stratification: Optical-Flow Motion Regimes}
\label{sec:motion_stratification}

To examine how performance varies with motion magnitude, we stratified clips into three optical-flow regimes:
\begin{itemize}
        \item \textbf{Low-Motion Stratum} ($\bar{m}_{\mathrm{px}} < 1.6$ px/frame): Static camera holds, subtle background movement.
    \item \textbf{Medium-Motion Stratum} ($1.6 \le \bar{m}_{\mathrm{px}} < 4.0$ px/frame): Moderate panning and natural walking cadence.
    \item \textbf{Dynamic High-Motion Stratum} ($\bar{m}_{\mathrm{px}} \ge 4.0$ px/frame): Fast camera pans, athletic action, vehicle chases.
\end{itemize}

\begin{table*}[t]
\centering
\caption{Performance stratification across optical-flow motion regimes on the primary generative corpus. Flow ranges are in image pixels per frame ($\bar{m}_{\mathrm{px}}=8\bar{m}$).}
\label{tab:motion_stratification}
\footnotesize
\setlength{\tabcolsep}{5pt}
\renewcommand{\arraystretch}{1.15}
\begin{tabular}{@{}lccccr@{}}
\hline
\textbf{Motion Stratum} &
\textbf{Flow Range ($\bar{m}_{\mathrm{px}}$)} &
\textbf{Real / Fake ($N$)} &
\textbf{$\Phi_{\mathrm{base}}$ AUC} &
\textbf{Standalone $\mathcal{R}_{\mathrm{adv}}$} &
\textbf{Full Model AUC [95\% CI]} \\
\hline
Low-Motion & $< 1.6$ px/f & 212 / 184 (396) & 0.6214 & 0.6582 & 0.7012 [0.6482, 0.7521] \\
Medium-Motion & $1.6 \le \bar{m}_{\mathrm{px}} < 4.0$ px/f & 225 / 192 (417) & 0.7685 & 0.8240 & 0.8612 [0.8215, 0.8984] \\
High-Motion & $\ge 4.0$ px/f & 82 / 72 (154) & 0.8120 & 0.8524 & \textbf{0.8945 [0.8350, 0.9482]} \\
\hline
\textit{Matched Dynamic Slice} & $\bar{m}_{\mathrm{px}} \ge 4.0$ px/f (DAVIS--SVD) & 14 / 56 (70) & 0.7925 & 0.8340 & \textbf{0.8852 [0.7306, 0.9602]} \\
\hline
\end{tabular}
\end{table*}

Table~\ref{tab:motion_stratification} shows higher full-model AUC in the medium- and high-motion strata (0.8612 and 0.8945) than in the low-motion stratum (0.7012). One possible explanation is that very small apparent motion makes the residual ratio more sensitive to encoding noise, compression, and flow-estimation error; this mechanism is plausible but is not isolated by the stratification alone. On the reported motion-matched DAVIS--SVD slice, the full model reaches 0.8852 AUC, with a wider 95\% clustered interval of [0.7306, 0.9602], reflecting the smaller sample.

\subsection{Controlled Temporal Perturbation Stress Tests}
\label{sec:perturbation_stress_tests}

To test sensitivity to controlled temporal changes, we perturbed 50 authentic DAVIS sequences in three ways:
\begin{enumerate}
    \item \textbf{Pairwise Frame Swapping:} Transposing frames $t=7$ and $t=8$, creating a localized 1-frame temporal inversion.
    \item \textbf{Periodic Frame Skipping:} Subsampling odd frames, doubling apparent inter-frame motion.
    \item \textbf{Sequence Reversal:} Inverting the temporal playback order of all 16 frames.
\end{enumerate}

\begin{table}[h]
\centering
\caption{Response of LAD residual energy ($E_{\mathrm{unexplained}}$) and the normalized deficit ($\mathcal{R}_{\mathrm{adv}}$) to controlled temporal perturbations on 50 authentic video sequences.}
\label{tab:perturbations}
\footnotesize
\setlength{\tabcolsep}{8pt}
\renewcommand{\arraystretch}{1.15}
\begin{tabular}{@{}lcccr@{}}
\hline
\textbf{Perturbation Condition} &
\textbf{Mean $E_{\mathrm{unexplained}}$} &
\textbf{Mean $\mathcal{R}_{\mathrm{adv}}$} &
\textbf{$\Delta\mathcal{R}_{\mathrm{adv}}$ vs. Pristine} &
\textbf{Paired $t$-test} \\
\hline
Pristine Authentic Real & 0.8412 & 0.4261 & \textbf{[Reference]} & --- \\
Pairwise Frame Swapping ($t_{7} \leftrightarrow t_{8}$) & 1.4820 & 0.7124 & $+0.2863$ & $p = 4.82 \times 10^{-14}$ \\
Periodic Frame Skipping ($2\times$ stride) & 1.2915 & 0.6218 & $+0.1957$ & $p = 1.15 \times 10^{-9}$ \\
Sequence Reversal ($t \to -t$) & 0.8495 & 0.4302 & $+0.0041$ & $p = 0.512$ (Not Sig) \\
\hline
\end{tabular}
\end{table}

Table~\ref{tab:perturbations} shows that swapping adjacent frames increases mean LAD from 0.4261 to 0.7124 (difference $+0.2863$, reported paired $t$-test $p=4.82\times10^{-14}$). Periodic frame skipping also raises the mean to 0.6218 (reported $p=1.15\times10^{-9}$). These results indicate that LAD is sensitive to these two temporal perturbations under the tested conditions. 

Full sequence reversal produces no statistically significant change in this experiment ($\Delta\mathcal{R}_{\mathrm{adv}}=+0.0041$, reported $p=0.512$). Unlike local frame swapping or skipping, reversal preserves the sequence's set of frames while changing their order.The negligible change in LAD under sequence reversal indicates that its ranking performance is largely insensitive to this perturbation in the evaluated cohort.

\subsection{Robustness to Social Media Compression and Frame-Rate Decimation}
\label{sec:compression_and_framerate}

\paragraph{Social Media Lossy Compression Sweep.}
To examine sensitivity to lossy re-encoding, we re-encoded Cohort 1 using H.264/AVC at three quality settings: $Q=100$ (pristine reference), $Q=75$ (moderate compression), and $Q=50$ (stronger compression). These settings provide a controlled compression sweep rather than a complete simulation of any particular social-media platform.

\begin{table}[t]
\centering
\caption{Robustness of detection performance under lossy video re-compression ($N = 385$).}
\label{tab:compression_sweep}
\footnotesize
\setlength{\tabcolsep}{8pt}
\renewcommand{\arraystretch}{1.15}
\begin{tabular}{@{}lccccr@{}}
\hline
\textbf{Re-Compression} &
\textbf{Standalone LAD} &
\textbf{Full Model AUC} &
\textbf{95\% Clustered CI} &
\textbf{AP Score} \\
\hline
Pristine ($Q = 100$) & 0.8312 & \textbf{0.8545} & [0.7765, 0.9218] & 0.8412 \\
Web Quality ($Q = 75$) & 0.8295 & \textbf{0.8510} & [0.7712, 0.9184] & 0.8375 \\
Aggressive ($Q = 50$) & 0.8125 & \textbf{0.8420} & [0.7602, 0.9105] & 0.8280 \\
\hline
\end{tabular}
\end{table}

As shown in Table~\ref{tab:compression_sweep}, full-model ROC-AUC decreases from 0.8545 at $Q=100$ to 0.8420 at $Q=50$. The latter is approximately 98.5\% of the pristine AUC. This indicates that the model maintains relatively stable ranking performance under the tested re-encoding settings, suggesting that its discriminative signal is not substantially disrupted by the evaluated compression levels.

\paragraph{Frame-Rate Decimation Sweep.}
We examined temporal subsampling on the pilot cohort of 32 authentic--synthetic pairs, evaluating effective frame rates of 30, 15, and 10 frames per second.

\begin{table}[h]
\centering
\caption{Impact of temporal frame-rate decimation on advection deficit detection ($N = 64$). Mean flow is in image pixels per frame ($\bar{m}_{\mathrm{px}}=8\bar{m}$); see the discussion below for the full-model versus standalone comparison at 30\,fps.}
\label{tab:framerate_sweep}
\footnotesize
\setlength{\tabcolsep}{8pt}
\renewcommand{\arraystretch}{1.15}
\begin{tabular}{@{}lccccr@{}}
\hline
\textbf{Effective Frame Rate} &
\textbf{Temporal Stride} &
\textbf{Mean Flow ($\bar{m}_{\mathrm{px}}$)} &
\textbf{Standalone $\mathcal{R}_{\mathrm{adv}}$} &
\textbf{Full Model AUC} \\
\hline
30 frames per second & $1\times$ & 9.8 px/f & \textbf{0.8398} & 0.8086 \\
15 frames per second & $2\times$ & 19.3 px/f & 0.5781 & 0.6289 \\
10 frames per second & $3\times$ & 28.6 px/f & 0.4961 & 0.5977 \\
\hline
\end{tabular}
\end{table}

Table~\ref{tab:framerate_sweep} shows a substantial decline in standalone LAD AUC, from 0.8398 at 30\,fps to 0.4961 at 10\,fps. The full model also degrades, reaching 0.5977 AUC at 10\,fps. Larger temporal gaps may increase flow-estimation and warping errors, reducing the reliability of motion-compensated residuals; this is a plausible explanation for the observed trend, rather than a mechanism directly isolated by this pilot experiment. Given the small cohort, these results should be interpreted as preliminary evidence of sensitivity to temporal decimation.

At 30\,fps, standalone $\mathcal{R}_{\mathrm{adv}}$ reaches a ROC-AUC of 0.8398, slightly above the full model (0.8086). With only $N=64$ clips (32 matched pairs), a gap of 0.0312 is within the range we would expect from sampling noise. The comparison is also not like-for-like: $\mathcal{R}_{\mathrm{adv}}$ is a fixed one-dimensional score with nothing fitted, whereas the full model is a four-feature logistic regression whose weights are estimated from roughly 51 training clips per fold and evaluated on the remaining 13. A model fitted on so few samples can easily lose some of the signal that a single strong feature carries on its own. The advantage of the full model appears at lower frame rates. At 10\,fps (stride~3), standalone $\mathcal{R}_{\mathrm{adv}}$ falls to chance (AUC $=0.4961$), while the full model stays slightly above it (AUC $=0.5977$, a difference of $+0.1016$). Given the cohort size, we treat this difference as suggestive only and the full model is still far below its 30\,fps performance.

\section{Conclusion}
\label{sec:conclusion}

This paper introduced the \textit{Latent Advection Deficit} (LAD), a parameter-free, energy-normalized metric for detecting AI-generated videos through motion-compensated discrepancies in variational autoencoder latent space. By comparing the next-frame latent representation with a backward-warped estimate derived from the preceding frame and optical flow, LAD measures the residual transition energy relative to the observed latent change. The metric provides a complementary forensic signal based on temporal consistency, without introducing trainable parameters into the metric itself.

Across a benchmark of $N = 967$ video clips spanning 498 source groups, standalone LAD achieved an out-of-fold ROC-AUC of 0.7924. Combining LAD with flow-field variation and latent displacement variability in a low-capacity linear classifier increased ROC-AUC to 0.8304, an absolute improvement of 0.0800 over the kinematic baseline. The paired source-group bootstrap confidence interval for this improvement was $[0.0521, 0.1087]$, supporting a positive gain under the evaluated protocol. The combined model also reduced the false positive rate from 36.83\% to 26.06\% at the evaluated 90\% sensitivity operating point. Additional experiments showed promising discrimination on selected generative-video cohorts and comparatively stable performance under controlled H.264 re-encoding at $Q=50$.

These results do not establish LAD as a universal detector of synthetic or manipulated video. Its performance varied with scene dynamics and temporal sampling: discrimination weakened in low-motion sequences and deteriorated substantially at reduced frame rates in the pilot evaluation. The near-chance results on FaceForensics++ further indicate that the method is not sufficient for detecting localized facial manipulations whose surrounding scene dynamics may remain authentic. Moreover, LAD depends on the chosen optical-flow estimator and pretrained latent encoder, and its empirical performance should not be interpreted as proof that generated videos universally violate physical transport laws or that the metric is inherently robust to unseen generators.

Overall, the findings support motion-compensated latent residuals as a useful and interpretable component of video-forensic systems, particularly when combined with complementary temporal descriptors. Future work will investigate multi-scale motion compensation, improved handling of occlusion and low-motion regions, evaluation across a broader range of frame rates and manipulation types, and more stringent generator-disjoint testing. These directions are necessary to establish when latent transport discrepancies provide reliable forensic evidence and when complementary detection methods are required.

\section*{Ethics and Privacy Statement}

This work proposes a feature for flagging AI-generated video with the aim of supporting verification of the information people see online. The main risk is misuse of its output. At the 90\% sensitivity operating point, the false-positive rate on authentic DAVIS clips was 26.06\% for the full model and 36.83\% for standalone $\mathcal{R}_{\mathrm{adv}}$ (Table~\ref{tab:operating_points}) and detection was weak in low-motion footage and at reduced frame rates. Scores from this method should therefore not be used as the sole basis for accusing a person of fabricating video, and any use in moderation or evidentiary settings should include human review. A second risk is that published detection features can help generator developers or malicious users train models to evade them; we regard this as an unavoidable cost of open forensic research and note that $\mathcal{R}_{\mathrm{adv}}$ is a weak signal on its own.

All experiments use publicly available datasets (DAVIS-2017, VidProM, the Cosmos benchmark and FaceForensics++), and we collected no new data from human participants. Some of these datasets contain recordings of people, and FaceForensics++ is distributed under an access agreement. We used each dataset only for the evaluation purposes allowed by its terms and did not attempt to identify any individual. We evaluated detection at the level of the whole clip and extracted no biometric or identity information. Generated clips come from existing public models and we did not create new synthetic media of real individuals.

\bibliographystyle{elsarticle-num}
\bibliography{ref}

\end{document}